\pdfoutput=1
\documentclass[lettersize,journal]{IEEEtran}
\usepackage{amsmath,amsfonts}
\usepackage{algorithmic}
\usepackage{algorithm}
\usepackage{array}
\usepackage[caption=false,font=normalsize,labelfont=sf,textfont=sf]{subfig}
\usepackage{textcomp}
\usepackage{stfloats}
\usepackage{url}
\usepackage{verbatim}
\usepackage{graphicx}
\usepackage{cite}
\usepackage{multirow}
\usepackage{amssymb}
\usepackage{booktabs}

\begin{document}

\title{Free-Form Composition Networks for Egocentric Action Recognition}


\author{Haoran~Wang,
        Qinghua~Cheng,
        Baosheng~Yu,
        Yibing~Zhan,
        Dapeng~Tao,
        Liang~Ding,
        and Haibin~Ling 
\thanks{H. Wang and Q. Cheng are with College of Information Science and Engineering, Northeastern University, Shenyang 110819, China.
E-mail: wanghaoran@ise.neu.edu.cn 2100723@stu.neu.edu.cn.}
\thanks{B. Yu is with School of Computer Science,
Faculty of Engineering, the University of Sydney, Darlington, NSW 2008, Australia.
E-mail: baosheng.yu@sydney.edu.au.}
\thanks{Y. Zhan and L. Ding are with JD Explore Academy, Beijing 100176, China.
E-mail: zybjy@mail.ustc.edu.cn liangding.liam@gmail.com.}
\thanks{D. Tao is with School of Information Science and Engineering, Yunnan University, Kunming 650091, Yunnan, China.
E-mail: dapeng.tao@gmail.com.}
\thanks{H. Ling is with Department of Computer Science, Stony Brook University, Stony Brook, USA.
E-mail: hling@cs.stonybrook.edu.}
}
\markboth{Journal of \LaTeX\ Class Files,~Vol.~14, No.~8, August~2023}%
{Shell \MakeLowercase{\textit{et al.}}: A Sample Article Using IEEEtran.cls for IEEE Journals}


\maketitle

\begin{abstract}
Egocentric action recognition is gaining significant attention in the field of human action recognition. In this paper, we address data scarcity issue in egocentric action recognition from a compositional generalization perspective. To tackle this problem, we propose a free-form composition network (FFCN) that can simultaneously learn disentangled verb, preposition, and noun representations, and then use them to compose new samples in the feature space for rare classes of action videos. First, we use a graph to capture the spatial-temporal relations among different hand/object instances in each action video. We thus decompose each action into a set of verb and preposition spatial-temporal representations using the edge features in the graph. The temporal decomposition extracts verb and preposition representations from different video frames, while the spatial decomposition adaptively learns verb and preposition representations from action-related instances in each frame. With these spatial-temporal representations of verbs and prepositions, we can compose new samples for those rare classes in a free-form manner, which is not restricted to a rigid form of a verb and a noun. The proposed FFCN can directly generate new training data samples for rare classes, hence significantly improve action recognition performance. We evaluated our method on three popular egocentric action recognition datasets, Something-Something V2, H2O, and EPIC-KITCHENS-100, and the experimental results demonstrate the effectiveness of the proposed method for handling data scarcity problems, including long-tailed and few-shot egocentric action recognition.
\end{abstract}


\section{Introduction}
\label{sec:intro}

\begin{figure}[!ht]
  \centering
  \includegraphics[width=\linewidth]{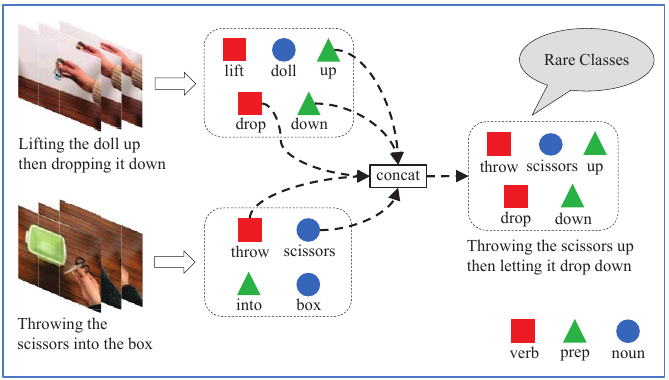}
  \caption{An illustration of the free-form compositional learning for egocentric action recognition. Existing compositional models usually decompose and compose action samples in a rigid combination, e.g., a verb and a noun, which is not suitable for the complex action analysis. In contrast, we extract the features of multiple verbs, prepositions, and nouns appearing in the action class, and then generate new training data for the rare actions in a relatively free-form manner.}
  \label{fig:motivation}
\end{figure}

\IEEEPARstart{E}{gocentric} action recognition has been attracting increasing amount of attention due to its wide range of applications containing robotics, augmented reality, and surveillance.  Different from the third-person view, egocentric video recognition mainly aims to identify the fine-grained hand-object interactions at a close distance, where the actions are usually described by a simple combination of a verb and a noun~\cite{damen2022rescaling,kwon2021h2o}. Following such a formulation, two-stream architectures have been widely used to separately classify the action and the active object with deep neural networks~\cite{ma2016going} and  are further improved by using either hand/object features~\cite{wang2021interactive,tekin2019hO,garcia2018first} or attention mechanisms~\cite{wang2020symbiotic,WangWZY2020AAAI}. However, existing methods mainly rely on a large amount of labelled data for training, leaving the data scarcity problem, such as long-tailed, few-shot, and zero-shot classifications, in egocentric action recognition under explored.

Data augmentation methods have been intensively explored for improving the variousness of training samples and preventing overfitting to a certain extent, especially in modern deep learning models. Vanilla augmentation strategies usually perform a set of operations on the original data, such as random rotation and random crop, while more advanced methods have been also proposed for the rare classes to augment the samples in multiple tasks such as semantic segmentation~\cite{yuan2021simple,zang2021fasa}, image classification~\cite{chubll2020feature,chen2019image}, and face recognition~\cite{yin2019feature,liu2020deep}.
Currently, compositional learning frameworks have proven effective in addressing the challenge of data scarcity in the detection of human-object interactions~\cite{HouYQPT2021CVPR,HuynhE2021iccv,HouYuQiao2021CVPR}. In these frameworks, all human-object interactions are broken down and reassembled to create new training samples, typically represented as a verb-noun pair. However, compositional action recognition remains relatively unexplored due to the inherent complexities involved in decomposing and composing action videos that contain both spatial and temporal semantic features~\cite{materzynska2020something}.
To elaborate further, on one hand, it is challenging to pinpoint the specific spatial and temporal regions within a video that correspond to the verb associated with a particular action class. On the other hand, action videos tend to convey a much richer set of semantic information compared to still images, making it difficult to encapsulate human actions within a rigid template, such as a verb and a noun. These inherent challenges have hindered the exploration of compositional action recognition techniques.
To address these issues, as shown in Figure~\ref{fig:motivation}, we learn the visual features of all the verbs, prepositions, and nouns from two actions ``lifting the doll up then dropping it down" and ``throwing the scissors into the box", and then compose new samples of ``throwing the scissors up then letting it drop down" with above verbs, prepositions, and nouns. Notably, compared with existing compositional models~\cite{Kim2022CVPR,wang2021interactive,HouYQPT2021CVPR}, rather than decomposing and composing action samples in a rigid combination of a verb and a noun, we simultaneously extract the features of multiple verbs, prepositions, and nouns from egocentric action videos, and then generate new training data in a relatively free-form manner for the rare classes.

To address the data scarcity problem in egocentric action recognition via compositional learning, we devise the \textit{free-form compositional networks} (FFCN). Compared with traditional compositional learning models, FFCN decomposes/composes human actions with much richer semantic information in a flexible manner. Specifically, it represents each hand-object interaction video by a graph and extracts the semantic features (i.e., verbs, prepositions, and nouns) from action videos. Given the locations of hands and objects, we utilize the temporal dynamics of hand-object and object-object relationships with the graph edge features to represent verbs and prepositions. Since it is common that each action description may simultaneously contain multiple verbs and prepositions, we propose to decompose the graph model to separately extract these spatial-temporal verb/preposition representations. The temporal decomposition aims to distinguish the verbs/prepositions appearing in different video segments, while the spatial decomposition adaptively learns discriminative sub-graphs to characterize verbs and prepositions. After that, we can utilize popular CNNs to extract object features (nouns), and then shuffle these verb, preposition, and noun representations to compose new samples for the rare action classes in the feature space.

The main contributions of this paper can be summarized as follows:
\begin{itemize}
\item We propose a new framework, referred to as the \textit{free-form compositional networks} (FFCN), to simultaneously extract the feature representations of multiple verbs and prepositions appearing in the action description based on the hand/object locations;
\item We explore the proposed method for generating new training samples with verb, preposition, and noun representations in a flexible free-form manner to alleviate the data scarcity problem (e.g., long-tailed and few-shot) in egocentric action recognition;
\item We intensively test our FFCN for egocentric action recognition in multiple data scarcity tasks, and the experimental results show the effectiveness of the proposed method for improving the model generalizability.
\end{itemize}

\begin{figure*}
  \centering
  \includegraphics[width=1.0\linewidth]{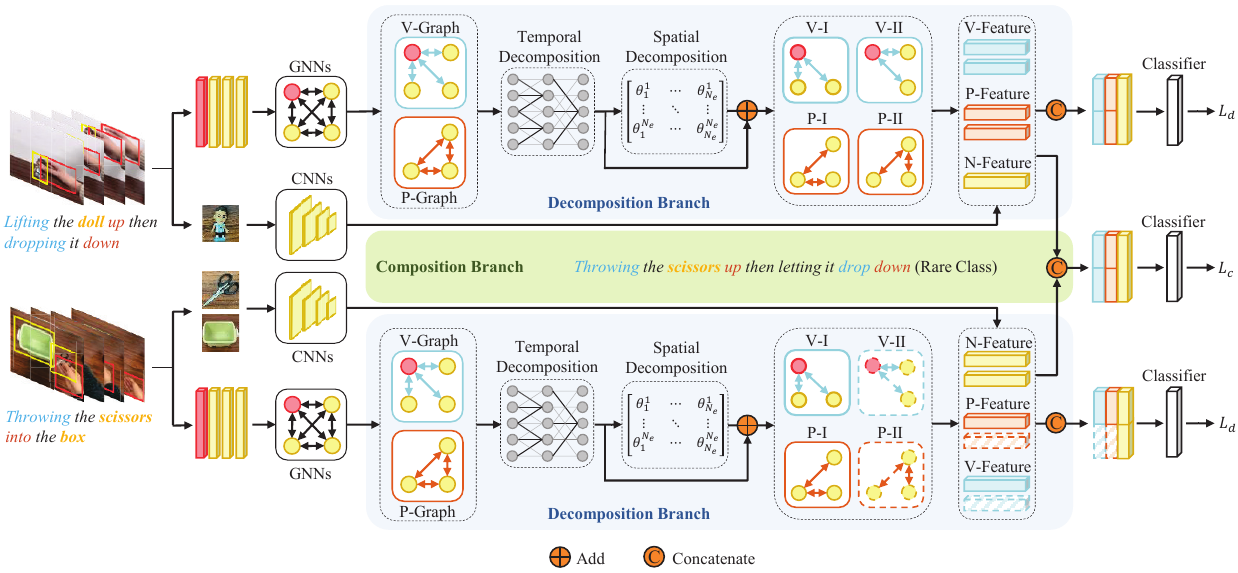}
  \caption{The main structure of the \textit{Free-Form Compositional Networks} (FFCN) for egocentric action recognition. We first build a graph to model the spatial and temporal interactive relations between instances appearing in an action video. In terms of the human-object and object-object relations, the whole graph is then disentangled into a verb graph (V-Graph) and a preposition graph (P-Graph) via the edge features in GNNs. After that, we temporally and spatially decompose a set of verb and preposition graphs across all the video frames to extract the features of all the verbs (e.g., V-I and V-II) and prepositions (e.g., P-I and P-II)  appearing in the action description. In addition, we extract the noun features by CNNs from the object regions in the video. Lastly, the above verb, preposition, and noun features can be combined in a flexible manner to generate new action samples for the rare action classes.}
  \label{fig:framework}
\end{figure*}

\section{Related Work}
\label{section:RelatedWork}

\subsection{Egocentric Action Recognition}

Compared with third-person views, egocentric action recognition relies on modeling the fine-grained hand-object interactions. Early methods mainly explore motion cues to distinguish egocentric actions. In particular, the optical flow-based global~\cite{KitaniOSS2011CVPR} and local~\cite{ryoo2013first} descriptors are designed to classify a variety of interactive actions. The motion descriptors are further combined with CNN-based appearance~\cite{ryoo2015pooled,poleg2016compact} and depth~\cite{tang2017action} information to improve the action representations. Noticing that the attention mechanism is efficient to locate the region of interest from cluttered background, the gaze estimation~\cite{li2018eye,li2015delving} is utilized to direct deep learning models to concentrate on the informative region in hand-object interaction videos. LSTA~\cite{sudhakaran2019lsta} extends traditional LSTM with the spatial attention based neural unit in order to concentrate on the action-related portion of the video sequence. In addition, the visual and audio features are fused by multi-scale temporal-binding~\cite{kazakos2019epic} to obtain complementary information from egocentric actions.

According to the structure of egocentric actions, recent methods utilize two independent branches to separately recognize the associated verb and noun. The twin stream network~\cite{ma2016going} utilizes the hand segmentation to locate the action-related object, and then classifies the verb and noun by CNNs. To exploit the correlation between the verb and noun branches, the symbiotic attention~\cite{wang2020symbiotic,WangWZY2020AAAI} uses the spatial locations and discriminative object features to focus on the occurring interactions. Based on the multi-task framework, H+O~\cite{tekin2019hO} estimates the hand and object poses in the 3D space from 2D RGB images, and then models the hand-object relationships with LSTM to recognize the action class and the action-related object. Several datasets~\cite{kwon2021h2o,garcia2018first} also provide annotated 3D hand poses and 6D object poses to facilitate the study of egocentric actions. To avoid labor-intensive annotations, IPL~\cite{wang2021interactive} leverages the motion cues to learn the verb class, and then guide the noun classification by extracting the action-related object. Compared with the rigid combination of a verb and a noun, our method is able to simultaneously extract the features of multiple verbs, prepositions, and nouns from an egocentric action video, and then reassemble these features to generate new training data for the rare and unseen actions.

\subsection{Compositional Learning}

Compositional learning has attracted increasing attention by synthesizing new samples based on the representations of partial images~\cite{alfassy2019laso,misra2017red,tulsiani2017learning}. Compositional models are also utilized to ameliorate the detection of partly occluded objects with non-occluded portions~\cite{wang2020robust,kortylewski2020compo}. For visual question answering~\cite{andreas2016neural,hu2017learning}, the questions are first decomposed into syntactic sub-modules, which are then regrouped by question-related compositional networks.

Currently, a variety of compositional models are proposed to focus on the data scarcity issue in human-object interaction (HOI) detection. External lexical knowledge~\cite{kato2018compositional} is embedded into graph convolutional networks to generate new action-object pairs for zero-shot HOI. The target objects in images are also replaced with similar intra-class instances in other images to improve the diversity of training data~\cite{FangXSLL2021aaai}. IDN~\cite{LiLWLL2020nips} and VCL~\cite{hou2020visual} extract shared action and object features from HOI images via the transformation function in the multi-branch network, and then regroup these action and object features to compose novel HOI samples in the feature space. In order to detect rare and unseen categories in the open world, ATL~\cite{hou2021affordance} and FCL~\cite{hou2021detecting} combine verb representations and novel object representations to generate large-scale HOI samples. ICompass~\cite{huynh2021interaction} and GCA~\cite{knyazev2021generative} further explore multiple verb-noun pairs from a single image via cross-attention mechanism and generative adversarial networks. Compared with existing HOI models decomposing/composing still images with the rigid template of a verb and a noun, we devise a flexible form to decompose/compose egocentric action videos with multiple verbs, prepositions, and nouns.

\subsection{Graphical Models}

Graphical models are widely used to reason the long range relationships in multiple computer vision tasks, including semantic segmentation~\cite{xie2021scale,wang2021end}, scene graph generation~\cite{knyazev2021generative,zhang2019graphical}, and group activity recognition~\cite{deng2016structure}. By performing convolutional operations on the graph-structured data, graph convolutional networks (GCNs)~\cite{kipf2016semi} are adopted to learn the temporal evolution of human body skeletons. ST-GCN~\cite{yan2018spatial} adopts the spatial and temporal convolutions to model the evolution of human poses. 2s-AGCN~\cite{shi2019two} increases the flexibility of graph models and learn both the static and dynamic information from skeleton sequence.
Besides skeleton data, GCNs are also employed to model the relationships between object proposals extracted from the action video. STRG~\cite{wang2018videos} constructs the space-time region graph with object bounding boxes to reason the human-object and object-object relationships. LSTR~\cite{li2019long} takes 3D tubelets as graph nodes, and models the long-term temporal dynamics via GCNs to localize human actions.

Compared with GCNs operating on a graph with its edge fixed as a similarity scalar, graph neural networks (GNNs) updates the node and edge features alternatively by message passing procedure. dNRI~\cite{graber2020dynamic} predicts the states of the human body joints and their relations at every point in time based on GNNs and all the previous states. Similarly, ~\cite{yang2021learning} makes use of historical pose tracklets to predict the human poses in the following frame, which serves as a robust estimation in challenging scenarios. In addition to the human pose estimation, GNN-based models are combined with the pose-graph optimization to deal with the problem of camera pose estimation~\cite{li2021pogo}. Different from above graphical models operating on the complete graph, MUSLE~\cite{li2021representing} explores compact sub-graphs to capture discriminative patterns of human actions for classification. In this paper, we adopt the edge features in GNNs to characterize the verbs and prepositions, and then distinguish different verbs/prepositions by learning from corresponding sub-graphs.

\begin{figure*}[!ht]
  \centering
  \includegraphics[width=0.75\linewidth]{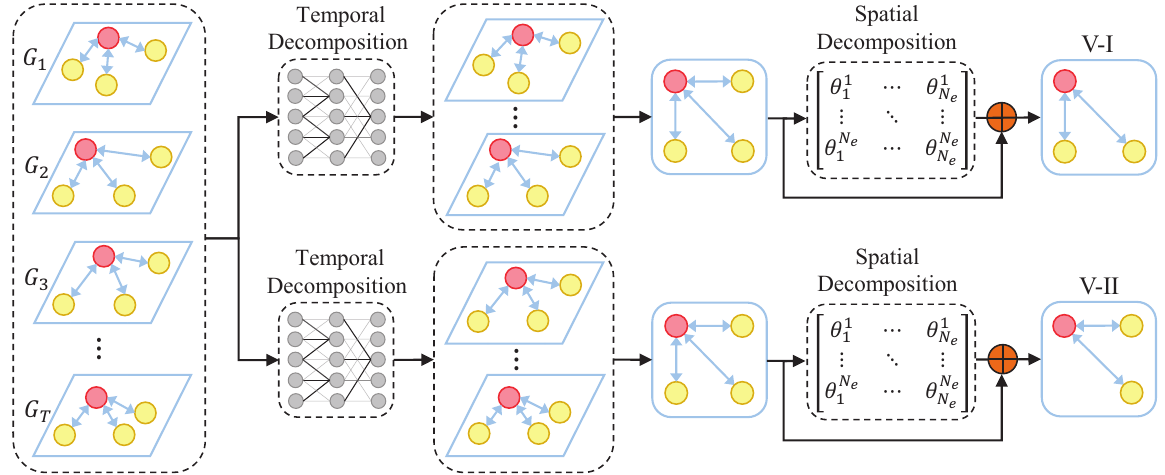}
  \caption{The procedure to extract the features of different verbs from a sequence of verb graphs constructed from $T$ video frames. The temporal decomposition module is first utilized to separate the video frames corresponding to different verbs. The verb graphs are then aggregated along the time dimension into a single graph model. After that, the spatial decomposition module is used to adaptively learn the verb feature from action-related instances. In the same way, the features of different prepositions can be extracted from the sequence of preposition graphs.}
  \label{fig:decompositon}
\end{figure*}

\section{Our Method}
\label{section:method}

In this section, we will begin by presenting an overview of the primary framework for egocentric action recognition, known as the Free-Form Composition Networks. Subsequently, we will delve into a comprehensive explanation of our proposed approach, which encompasses the spatial-temporal action decomposition modules. Lastly, we will outline the process of creating new action representations in a free-form manner.

\subsection{Overview}


We present an overview of the primary framework of the Free-Form Composition Networks (FFCN) in Figure~\ref{fig:framework}. This framework is designed to combine semantic elements from various videos to generate new instances corresponding to specific action classes. To accomplish this, we start by considering each video frame, which contains $N$ instances, including hands and objects, and establish a graph with $N$ nodes to represent these instances (the red and yellow nodes separately represent hands and objects). Subsequently, we use Graph Neural Networks (GNNs) to concurrently enhance the features of both nodes and edges in the graph. The relationships between verbs, prepositions, hands, and objects are closely intertwined within the edge features of the graph, which capture substantial relational information among different nodes. We then partition the graph into two sub-graphs, namely, a verb graph and a preposition graph, to extract the representations of verbs and prepositions from the edge features. Recognizing that action descriptions may not always comprise just one verb and one preposition, we introduce temporal and spatial decomposition modules to distinguish multiple verb and preposition features from a sequence of verb and preposition graphs.

The detailed procedure to extract different verb representations is shown in Figure~\ref{fig:decompositon}. Since the number of verbs varies between different action classes, we consistently extract ${n_v}$ verb representations from each video, where ${n_v}$ indicates the largest number of verbs appearing in an action description. Similarly, the preposition representations are extracted from the sequence of preposition graphs. To extract the object representations corresponding to the nouns in the action description, we employ vanilla CNNs. With the above-mentioned shared verb/preposition/noun representations, we then integrate those verb, preposition, and noun representations across different classes to generate new training samples in the composition branch, with a focus on the rare classes. That is, the verb, preposition, and noun representations are concatenated in terms of their order appearing in the action description.


It is important to note that the quantity of semantic components (verbs, prepositions, and nouns) varies significantly across different action classes. For the rare classes, which may contain ${n_v}$ verbs, ${n_p}$ prepositions, and ${n_n}$ nouns, we take the $d$-dimensional features corresponding to verbs, prepositions, and nouns and individually reduce their dimensions to $d/{n_v}$, $d/{n_p}$, and $d/{n_n}$ using a multilayer perceptron (MLP). After this dimension reduction, these features are then concatenated based on their order of appearance in the action description, resulting in the creation of new samples with a dimension of $3d$. This ensures that the features of all generated training samples are normalized to the same dimension.

\subsection{Graph Generation}

In this subsection, we define the process of generating a graph as follows: Consider a video comprising $T$ frames, each containing $N$ instances. We represent all these instances' feature information as $X = (x_1^1, \cdots ,x_N^1,x_1^2, \cdots ,x_N^2, \cdots ,x_1^T, \cdots ,x_N^T)$, where $x_i^t$ represents the feature representation of the $i$-th instance in frame $t$. These instance representations can take various forms, such as 1) bounding box coordinates that include the center position, height, and width of each bounding box, and 2) appearance data obtained from the backbone CNNs. In this study, we choose to use the bounding box coordinates as the instance features for extracting verb and preposition characteristics. This choice is motivated by two reasons: firstly, these coordinates complement the widely adopted RGB-based methods, and secondly, verbs and prepositions have a closer association with the spatial positions of hands and objects than with RGB information.
For example, two action videos of ``pour juice" and ``pour water" contain the same verb ``pour", but they may have different appearance features. The zero vectors are adopted as the features when there are fewer instances detected in each video frame.
Therefore, we have the ability to build a graph denoted as $G = (V,E)$ where the graph nodes and edges symbolize the identified instances and their interconnections. To enhance this graph, we employ a graph neural network (GNN). In the process of message passing at step $s$, we represent the node features as $h_{i,t}^s$ and the edge features as $f_{ij,t}^s$. The architecture of the GNN, which concurrently improves both the graph nodes and edges, is structured as follows:
\begin{eqnarray}
\label{eq:embedini}
h_{i,t}^0 &=& {\phi _{\rm ini}}(x_i^t),
\\
\label{eq:embededge}
f_{ij,t}^s &=& {\phi _{\rm edge}} ([h_{i,t}^{s-1},h_{j,t}^{s-1}]),~~
\\
\label{eq:embednode}
h_{i,t}^s &=& {\phi_{\rm node}} (\sum\limits_{j \ne i} {f_{ij,t}^s} ).
\end{eqnarray}
Here, the notation $[ \cdot , \cdot ]$ signifies the concatenation operation, while $h$ and $f$ correspond to the intermediate hidden states of the graph node and edge, respectively. Additionally, $\phi$ stands for the multilayer perceptrons (MLPs).


We utilize the output of the GNNs, which are the graph edge features, to gain insights into the interactive relationships among various instances. Recognizing that the verb and preposition characteristics can be respectively attributed to the hand-object and object-object relationships within egocentric videos, we proceed to break down the refined graph model into two distinct components: the verb graph and the preposition graph. This allows us to independently capture the features related to the verb and preposition.
Within the verb graph, we establish connections between the hand node and all the object nodes to extract information regarding hand-object relationships. In contrast, in the preposition graph, we interconnect the object nodes themselves to capture the interactions among objects. This approach differs from most existing graph-based models, which predominantly focus on exploring the features of graph nodes. In our case, we leverage the robust edge features to extract semantic components, namely the verb and preposition features.

\subsection{Temporal Decomposition}

In this subsection, we formulate the temporal decomposition as follows. Given the fact that the action description usually consists of multiple verbs and prepositions, which correspond to different segments in the video. Meanwhile, the dilated convolution is usually adopted to aggregate multi-scale contextual information, and increase the receptive field by stacking multiple layers. Therefore, we propose to explore dilated convolutions on the verb/preposition graph edges in the time dimension to extract the features of different verbs/prepositions from the corresponding video segments.

Given ${N_e}$ edges in the verb graph, the features of the $k$\textit{-th} edge across $T$ frames are denoted as ${F_k} = ({f_{k1}},{f_{k2}},\cdots,{f_{kT}})$, which are fed into several layers of dilated 1D convolutions. In this paper, we use four layers of dilated convolutions with dilation factors $d = 1,2,3,4$ at each layer. All these layers have the same number of convolutional filters with the kernel size 3. Each layer utilizes a dilated convolution with the ReLU activation function, and the output is refined by the next layer. Inspired by MS-TCN~\cite{farha2019ms}, we also employ a residual connection~\cite{he2016deep} and the operations at each layer are formulated as follows:
\begin{eqnarray}
\label{eq:dilatedconv}
{\hat H_l} = \text{ReLU}({W_1} * {H_{l - 1}} + {b_1}),
\\
\label{eq:residual}
{H_l} = {H_{l - 1}} + {W_2} * {{\hat H}_l} + {b_2},~~
\end{eqnarray}
where ${H_l}$ indicates the output of the $l$\textit{-th} layer, $*$ is the convolution operator, ${W_1} \in {R^{3 \times D \times D}}$ are the weights of the dilated convolution with $D$ filters and kernel size 3, ${W_2} \in {R^{1 \times D \times D}}$ are the weights of the $1 \times 1$ convolution, and ${b_1},{b_2} \in {R^D}$ indicate the bias. The optimized graph edge features can be obtained from the last dilated convolution layer and the features of the $k$\textit{-th} edge across $T$ frames are then aggregated as follows:
\begin{eqnarray}
\label{eq:edgefea}
{q_k} = {\phi _{verb}}({f_{k1}}, \cdots ,{f_{kT}}),
\end{eqnarray}
where the function ${\phi _{verb}}$ indicates the MLP networks. All the edges in the verb graph share the same dilated temporal convolution layers, in order to extract the verb feature from the graph edges belonging to the same video segment. This way, all the verb graphs across $T$ frames are aggregated along the time dimension into a single graph model, where the edge features are denoted as ${Q_v} = [{q_1},{q_2}, \cdots ,{q_{{N_e}}}]$. Similarly, the preposition graphs across $T$ frames are temporally aggregated into a single graph with the edge features ${Q_p}$.

\subsection{Spatial Decomposition}

In this subsection, we formulate the spatial decomposition as follows.  Though the verb and preposition graphs are utilized to explore the verb and preposition features based on the temporal evolution of hand-object and object-object relationships, the verb and preposition graphs usually contain irrelevant objects due to the complex background, which may degrades the reasoning from spatial-temporal relationships. Instead of using a simple averaging pooling operation to aggregate the graph edge features, we propose to adaptively learn from the relationships between action-related instances to extract the verb and preposition features.

For the verb features, we apply the graph convolutional networks (GCNs) to encode the interactive relations between different instances as follows:
\begin{eqnarray}
\label{eq:gcn}
Z = A{Q_v^T}W,
\end{eqnarray}
where $A$ represents the adjacency matrix with ${N_e} \times {N_e}$ dimensions, ${Q_v}$ indicates the edge features in the verb graph with $D \times {N_e}$ dimensions, and $W$ is the weight matrix with the dimension $D \times D$. Following the same strategy in~\cite{shi2019two}, the elements $\theta$ in the adjacency matrix $A$ are trainable parameters in the networks and updated in the training process. Therefore, the output of each graph convolution layer $Z$ is in ${N_e} \times D$ dimensions, and multiple convolution layers can be stacked together. After the message passing within the verb graph, the final outputs of the GCNs then indicate the optimized relationships between different instances, which are aggregated by average pooling to obtain the verb feature. Notably, we always adopt a residual connection to facilitate the gradient flow. We can also extract the features of all the verbs and prepositions appearing in the action description in the same way. In addition, we adopt vanilla CNNs to extract the object features in each video frame, and then fuse the features of the same object across all the frames by MLPs to represent the nouns.

\subsection{Action Composition}


In this subsection, we outline the process of creating new action samples as follows: We start with verb, preposition, and noun features, and from these, we generate features for new action samples, which are aimed at enriching the training data for the less common classes in the composition branch. To construct these samples for rare actions, we follow the sequence of verbs, prepositions, and nouns as they appear in the infrequent action descriptions. We concatenate the corresponding features and employ MLPs to reduce the dimension of these features, effectively representing the rare action samples within the feature space. These generated samples are then used for training in conjunction with the original training samples. It's worth noting that both the composition and decomposition branches share the same weights in the fully connected and classifier layers. Finally, we simultaneously minimize two loss functions: ${L_d}$ for the decomposition branch and ${L_c}$ for the composition branch, expressed as $L = {L_d} + \lambda \cdot {L_c}$, where $\lambda \ge 0$ serves as a hyper-parameter to balance the two distinct branches.

\section{Experiments}
\label{section:Experiments}

In this section, we perform experiments on three popular datasets for egocentric action recognition, including Something-Something V2 (SS-V2)~\cite{goyal2017something},  H2O~\cite{kwon2021h2o}, and EPIC-KITCHENS-100 (KITCHENS)~\cite{damen2022rescaling}. Specifically, we first introduce the statistics of different datasets and implementation details on each dataset such as hyper-parameters.  We then compare the proposed method with recent state-of-the-art methods. Lastly, we perform comprehensive ablation studies to better understand how the proposed method addresses data scarcity issues for robust egocentric action recognition.

\subsection{Datasets}

\begin{itemize}

\item \textbf{Something-Something V2 (SS-V2)~\cite{goyal2017something}} is a large-scale hand-object interaction dataset, which contains 174 action categories created in a crowdsourcing manner. The videos in the same class are captured by performing the same action with various objects, in order to make the methods identify human actions regardless of the interactive objects. It provides the action descriptions mainly consisting of multiple verbs and prepositions as well as video annotations, including bounding boxes and their identities. Overall, it includes 168,913 training videos and 24,777 validation videos. The number of videos in different action classes follows a long-tailed distribution ranging from 115 to 4,081.

\item \textbf{H2O~\cite{kwon2021h2o}} is a recently introduced egocentric interaction dataset containing 36 action classes of two hands manipulating objects, which are captured by four participants in three different environments. The action labels are verb-noun pairs composed by 11 verb classes and 8 noun classes. This dataset includes 571,645 RGB-D frames with 344,645 training frames, 73,380 validation frames, and 153,620 testing frames. It also provides detailed annotations, containing hand and object poses separately in 3D and 6D space, object meshes, camera parameters, and point cloud data. We focus on the action recognition task, and provide the top-1 accuracy on the validation videos.

\item \textbf{EPIC-KITCHENS-100 (KITCHENS)~\cite{damen2022rescaling}} is a large-scale egocentric video dataset, which includes 100 hours of daily activity videos captured in kitchens. The performed activities are unscripted and close to real-world data, which makes the number of videos unbalanced between different classes. All the videos are split into train/validation/test with a ratio of 75/10/15. In total, there are 89,977 video clips of fine-grained interactions annotated with 97 verb classes and 300 noun classes. It also provides annotations of hands and objects appearing in the video. Following the standard evaluation strategy, we separately predict the verb and noun classes of each video, and then obtain the action class by fusing the verb and noun classification results. We provide the top-1 accuracy on the validation videos.
\end{itemize}

\subsection{Implementation Details}
\label{section:details}

We implement the proposed free-form composition networks (FFCN) based on PyTorch~\cite{paszke2019pytorch}, and sample 16 video frames from each clip. We adopt SGD~\cite{bottou2010large} with the momentum value of 0.9 and the weight decay value of 1e-4.

On the Something-Something V2 dataset, we respectively report the settings in the standard, object-independent, and few-shot classification tasks. In the standard and object-independent protocols, each mini-batch contains 64 samples. The basic learning rate is set to 0.05, and divided by a factor of 10 after 15 epochs. A total of 25 epochs are utilized to train the networks. We generate new training data for 30 tail classes, and compose 10 samples in each mini-batch. In the few-shot protocol, each mini-batch contains 16 samples and the learning rate is set to 0.05 with a total of 8 epochs. We compose new training data for all the classes, and the number of composed samples is equal to original training samples in each class. We set the weight $\lambda {\rm{ = }}0.1$ to balance two branches in the final loss function.

On the H2O dataset, each mini-batch contains 12 action samples. The basic learning rate is set to 0.005, and decreased by 0.1 at the 20\textit{-th} epoch with a total of 80 epochs. We generate new training data for each action class, and compose 10 action samples in each mini-batch. We utilize the 3D hand poses and 6D object poses provided by this dataset as the features of each instance to learn the verb features, and employ the VGG-16~\cite{simonyan2015very} to learn the noun features from the region of the objects. We set the weight $\lambda {\rm{ = }}0.2$ to balance the decomposition and composition branches.

On the EPIC-KITCHENS-100 dataset, each mini-batch contains 256 videos. The basic learning rate is set to 0.001, and divided by a factor of 10 at the 20\textit{-th} and 40\textit{-th} epochs. A total of 50 epochs are utilized to train the networks. We compose new samples for the tail classes with fewer than 100 training samples, and generate 20 samples in each mini-batch. We use the bounding box positions of hands and objects to learn the verb features, and adopt the VGG-16~\cite{simonyan2015very} to extract the noun features. We set the weight $\lambda {\rm{ = }}0.1$ in the final loss function.

\begin{table}[!ht]
\centering
  \begin{tabular}{clcc}
  \toprule
    & Method  &  Year   &  Accuracy \\
   \midrule
   \multirow{2}{*}{\rotatebox{0}{Loc.}}
   & STIN \cite{materzynska2020something}   & 2020  & $ 48.4\%$  \\
   & \textbf{FFCN}    & - & $\textbf{52.9\%}$  \\
  \midrule
  \multirow{10}{*}{\rotatebox{0}{RGB}}
  & I3D  ~\cite{carreira2017quo} &  2017  & $ 56.0\%$ \\
  & TRN Dual Atten.~\cite{xiao2019reasoning} &  2019  & $ 51.6\%$\\
  & TSM \cite{lin2019tsm}  & 2019  & $ 61.7\%$  \\
  & STIN + I3D \cite{materzynska2020something}   & 2020  &  $ 60.2\%$ \\
  & TDN \cite{wang2021tdn}  & 2021  &  $ 67.0\%$ \\
  & DirecFormer \cite{truong2022direcformer}  & 2022  &  $ 64.9\%$ \\
  & TCM \cite{liu2022motion}  & 2022  &  $ 67.8\%$ \\
  & SIFA \cite{long2022stand}  & 2022  &  $ 69.8\%$ \\
  & ViTTA \cite{LinMKPKB2023CVPR}  & 2023  &  $ 66.4\%$ \\
  & Video-FocalNet~\cite{wasim2023videofocalnets} & 2023 & $ 71.1\%$ \\
  & \textbf{FFCN} + I3D  & - & $ 67.3\%$ \\
  & \textbf{FFCN} + TDN  & - & $\textbf{74.8\%}$ \\
   \bottomrule
 \end{tabular}
 \vspace{2mm}
 \caption{Results on the original SS-V2 dataset}
 \label{tab:oristh}
 \end{table}

\begin{table}[!ht]
\centering
  \begin{tabular}{clcccc}
  \toprule
  & Method &  Year &  Obj-ind &  5-shot & 10-shot  \\
   \midrule
  \multirow{3}{*}{\rotatebox{0}{Loc.}}
  & STIN \cite{materzynska2020something} & 2020  & $ 51.3\%$ & $ 25.8\%$ & $ 32.9\%$ \\
  & STIN + NL \cite{materzynska2020something} & 2020  & $ 51.4\%$ & $ 27.7\%$ & $ 33.5\%$ \\
  & \textbf{FFCN}   & - & $\textbf{54.3\%}$ & $\textbf{30.9\%}$ & $\textbf{35.8\%}$ \\
  \midrule
  \multirow{9}{*}{\rotatebox{0}{RGB}}
  & I3D ~\cite{carreira2017quo}  & 2017  & $ 51.9\%$    &  $ 23.5\%$ & $ 27.1\%$ \\
  & STRG \cite{wang2018videos}                   & 2018  & $ 52.3\%$  &  $ 24.8\%$   &  $ 29.9\%$ \\
  & STIN + STRG \cite{materzynska2020something}  & 2020  & $ 56.2\%$  &  $ 29.1\%$   &  $ 34.6\%$ \\
  & STIN + I3D \cite{materzynska2020something}   & 2020  & $ 58.1\%$ & $ 34.0\%$  & $ 40.6\%$  \\
  & TDN \cite{wang2021tdn}                       & 2021  & $ 62.5\%$ & $ 30.8\%$  & $ 38.5\%$  \\
  & SIFA \cite{long2022stand}                    & 2022  & $ 44.6\%$  & $ 19.4\%$ & $ 23.7\%$ \\
  & TCM \cite{liu2022motion}                     & 2022  & $ 59.8\%$  & $ 33.6\%$ & $ 39.8\%$ \\
  & Video-FocalNet~\cite{wasim2023videofocalnets} & 2023 & $ 68.8\%$  & $ 19.3\%$ & $ 21.6\%$ \\
  & \textbf{FFCN} + I3D                           &  -    & $ 66.9\%$  & $ 38.7\%$ & $ 44.4\%$ \\
  & \textbf{FFCN} + TDN   & - & $\textbf{70.5\%}$ &  $\textbf{43.1\%}$   &  $\textbf{53.5\%}$ \\
   \bottomrule
\end{tabular}
\vspace{2mm}
\caption{Results of object-independent and few-shot action recognition on the SS-V2 dataset}
\label{tab:objfew}
\end{table}

\subsection{Results on Something-Something V2}


Initially, we evaluate our proposed method using the standard approach, which involves a dataset consisting of 168,913 training videos and 24,777 validation videos. The performance comparison against state-of-the-art methods is presented in Table~\ref{tab:oristh}. To ensure a fair assessment, we separately compare our Free-Form Composition Network (FFCN) with two different categories of existing methods that employ bounding box locations and video sequences as input to the networks. Both FFCN and STIN~\cite{materzynska2020something} utilize bounding box locations as input and can be integrated with RGB-based methods to enhance accuracy. In this scenario, FFCN outperforms STIN in both settings. Additionally, I3D~\cite{carreira2017quo} is a widely used RGB-based method for video action recognition. When combined with the same method, FFCN+I3D ($67.3\%$) surpasses STIN+I3D ($60.2\%$) by $7.1\%$. As TDN achieves better results than I3D, we incorporate FFCN with TDN, resulting in an accuracy of $74.8\%$, which is $7.8\%$ higher than TDN and notably superior to recently proposed alternatives.


Following the evaluation protocol outlined in~\cite{materzynska2020something}, we also assess the performance of our FFCN in an object-independent scenario. In this setup, the training set consists of combinations of nouns and verbs that do not appear in the testing set. The training and validation sets contain 54,919 and 57,876 videos, respectively, encompassing a total of 174 action classes. As demonstrated in Table~\ref{tab:objfew}, FFCN ($54.3\%$) outperforms STIN ($51.3\%$) by exclusively utilizing bounding box locations of hands and objects to capture the features of verbs and prepositions present in the action descriptions. Most existing methods rely on whole video frames as input to distinguish between different action classes, partially relying on scene appearance information rather than the motion characteristics between hands and objects. The proposed FFCN can be easily combined with RGB-based methods through simple late fusion, resulting in an improved performance. By fusing FFCN ($54.3\%$) with I3D ($51.9\%$), the final accuracy reaches $66.9\%$. When combined with TDN ($62.5\%$), we achieve the highest accuracy of all, at $70.5\%$.

\begin{figure*}
  \centering
  \includegraphics[width=\linewidth]{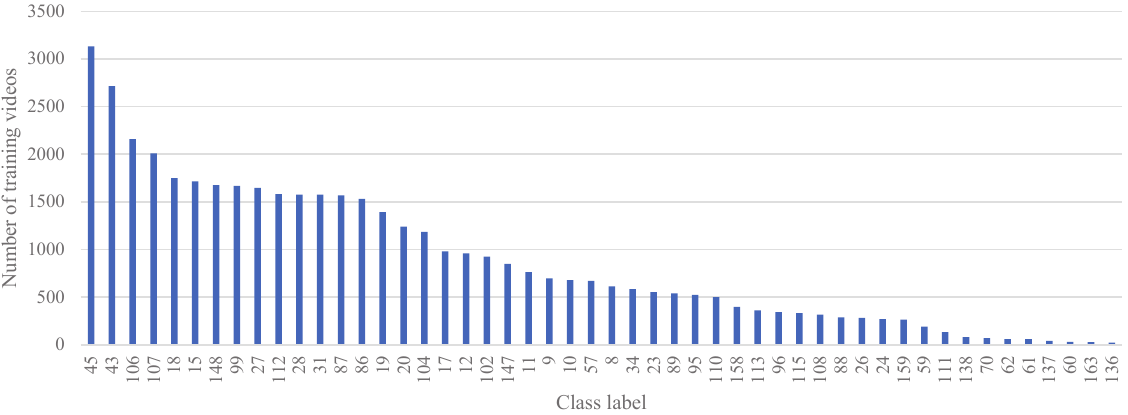}\\
  \caption{The selected action classes with shared verbs and prepositions from the Something-Something V2 dataset. The number of training samples in each class follows the long-tailed distribution.}
  \label{fig:seleclas}
\end{figure*}

To further validate the effectiveness of composing new training data, we test the performance of FFCN in the task of few-shot action recognition. We also utilize the protocol in~\cite{materzynska2020something} to separate the training data into the base and novel sets. Our networks are first trained on the 88 base classes, and then finetuned on the other 86 novel classes with 5 or 10 training videos in each class. As illustrated in Table~\ref{tab:objfew}, we separately compare the accuracies of 5-shot and 10-shot action recognition on the 86 novel classes. With only 5 training samples in each action class, our FFCN ($ 30.9\%$) exceeds STIN ($ 25.8\%$) by $ 5.1\%$ based on the locations of hands and objects. Combined with RGB-based models, FFCN+I3D ($ 38.7\%$) outperforms I3D ($ 23.5\%$) by $ 15.2\%$, and FFCN+TDN ($ 43.1\%$) outperforms TDN ($ 30.8\%$) by $ 12.3\%$. In the 10-shot action recognition, our FFCN also improves the performances of RGB-based models by more than $ 15\%$. The above results in the few-shot learning validate that our FFCN can augment the training data to obviously improve the accuracies in egocentric action recognition.

\subsection{Results on H2O}

\begin{table}
  \centering
  \begin{tabular}{lcc}
  \toprule
    Method   &  Year  &  Accuracy     \\
  \midrule
  I3D  ~\cite{carreira2017quo} &  2017  & $ 87.2\%$ \\
  ST-GCN \cite{yan2018spatial}   & 2018  & $ 83.5\%$  \\
  H+O \cite{tekin2019hO}   & 2019  & $ 80.5\%$  \\
  SlowFast~\cite{feichtenhofer2019slowfast} &  2019  & $ 86.0\%$ \\
  X-ViT~\cite{bulat2021space}  & 2021  & $ 82.8\%$  \\
  TA-GCN \cite{kwon2021h2o}   & 2021  & $ 86.8\%$  \\
  TAR~\cite{cho2022transformer} & 2022  &  $ 90.9\%$ \\
  HTT~\cite{Wen2023CVPR} & 2023  &  $ 86.4\%$ \\
  \midrule
  \textbf{FFCN}    & - & $\textbf{93.3\%}$  \\
  \textbf{FFCN} + X-ViT  & - & $\textbf{95.0\%}$ \\
  \textbf{FFCN} + I3D  & - & $\textbf{95.9\%}$ \\
   \bottomrule
 \end{tabular}
 \vspace{2mm}
 \caption{Results on the H2O dataset}
 \label{tab:h2o}
\end{table}

We also test our FFCN on the H2O dataset, and the comparison results are illustrated in Table~\ref{tab:h2o}. RGB-based methods are widely used to recognize hand-object interactions, such as I3D, SlowFast, and X-ViT. Moreover, TAR achieves excellent result ($ 90.9\%$) in ECCV 2022 challenge on action recognition from egocentric camera. HTT adopts the locations of 3D hand joints and the RGB features of objects together to obtain a promising result ($ 86.4\%$). The accuracy of our FFCN is $ 93.3\%$, which outperforms other methods by a clear margin with only location information. Because the location- and RGB-based methods are complementary in representing hand-object interactions, we combine our FFCN ($ 93.3\%$) with recently proposed X-ViT ($ 82.8\%$) to further increase the accuracy to $ 95.0\%$. We also fuse the results of FFCN and I3D, and the accuracy is improved to $ 95.9\%$, which is the best performance on the H2O dataset.

\subsection{Results on EPIC-KITCHENS-100}

Different from the above two datasets designed to directly predict the action labels, the EPIC-KITCHENS-100 dataset expects the methods to separately predict the verb, noun, and action labels of each video. Our FFCN mainly utilizes the coordinates of hands and objects to extract the motion features in the video without the appearance information, so we separately report the prediction accuracies of the verb and action in FFCN and the accuracies of the verb, noun, and action by integrating FFCN with a RGB-based method. The comparison results are illustrated in Table~\ref{tab:KITCHENS}. Because of the promising performance, SlowFast and TSM are both widely used to recognize egocentric actions. In first person videos, the active objects are usually surrounded by distracting objects, so IPL and SOS present two kinds of methods to capture the action-related objects in the video. Based on the transformer architecture, X-ViT exploits the spatial-temporal attention in action videos, and OMNIVORE jointly trains a single model from multiple modalities, such as images, videos, and 3D data, to achieve outstanding results in different classification tasks. All aforementioned methods are RGB-based models, and extract the video features from the appearance information. Our FFCN utilizes the locations of hands and objects to predict the verb and action labels, and the mean accuracies are separately $ 56.1\%$ and $ 52.2\%$. Integrated with X-ViT, the verb and action accuracies are increased to $ 71.5\%$ and $ 60.5\%$. Our FFCN is also fused with OMNIVORE, and the final verb and action prediction accuracies are $ 73.6\%$ and $ 69.1\%$, which are separately $ 4.1\%$ and $ 19.2\%$ higher than OMNIVORE.

\begin{table}
  \centering
  \begin{tabular}{lcccc}
  \toprule
    Method   &  Year  &  Verb  &  Noun   &  Action    \\
  \midrule
    SlowFast~\cite{feichtenhofer2019slowfast}&  2019  &  $ 65.6\%$ &  $ 50.0\% $ & $ 38.5\% $  \\
    TSM~\cite{lin2019tsm}                    &  2019  &  $ 67.9\%$  &  $ 49.1\% $ & $ 38.4\% $  \\
    IPL~\cite{wang2021interactive}           &  2021  &  $ 68.6\%$  &  $ 51.2\% $ & $ 41.0\% $  \\
    X-ViT~\cite{bulat2021space}              &  2021  &  $ 68.7\%$  &  $ 56.4\% $ & $ 44.3\% $  \\
    SOS~\cite{escorcia2022sos}               &  2022  &  $ 69.3\%$  &  $ 57.9\% $ & $ 45.7\% $  \\
    OMNIVORE~\cite{Girdhar2022CVPR}          &  2022  &  $ 69.5\%$  &  $ 61.7\% $ & $ 49.9\% $  \\
    Video-FocalNet~\cite{wasim2023videofocalnets} &  2023  &  $ 66.2\%$  &  $ 46.7\% $ & $ 40.9\% $  \\
    \midrule
    \textbf{FFCN}                            &   -    &  $ 56.1\%$  &      -      &  $ 52.2\% $ \\
    \textbf{FFCN}+X-ViT                      &   -    &  $ 71.5\%$  &  $ 56.4\% $ &  $ 60.5\% $  \\
    \textbf{FFCN}+OMNIVORE    &  -  &  $ \textbf{73.6\%}$  &  $ 61.7\% $ &  $ \textbf{69.1\%} $  \\
   \bottomrule
 \end{tabular}
 \vspace{2mm}
 \caption{Results on the EPIC-KITCHENS-100 dataset}
 \label{tab:KITCHENS}
\end{table}

\subsection{Ablation Studies}
\label{section:ablationstudy}


\begin{table}
  \centering
  \begin{tabular}{lcc}
  \toprule
    Dataset   & Decomp. branch & Two branches    \\
  \midrule
    SS-V2         &  $ 51.3\%$ & $ 52.9\%$   \\
    SS-V2-Subset  &  $ 55.1\%$ & $ 64.7\%$   \\
    H2O           &  $ 91.7\%$ & $ 93.3\%$   \\
    KITCHENS      &  $ 50.5\%$ & $ 52.2\%$   \\
   \bottomrule
 \end{tabular}
  \vspace{2mm}
 \caption{The effectiveness of decomposition and composition branches in the free-form composition networks}
 \label{tab:diffcomp}
\end{table}

\begin{figure*}
  \centering
  \includegraphics[width=1.0\linewidth]{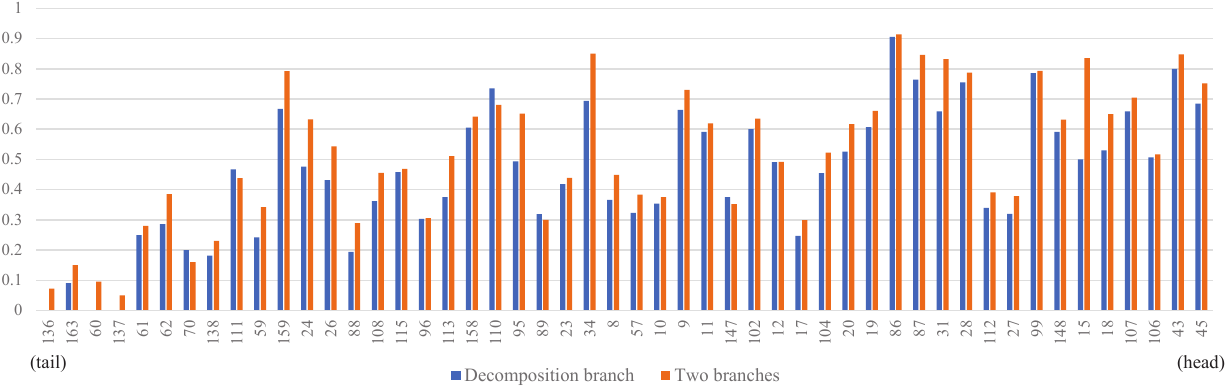}\\
  \caption{Results of each action class whether or not composing new training data on the SS-V2-Subset. After applying data augmentation, the accuracies of 45 classes are increased with 50 actions in total.}
  \label{fig:SScomp}
\end{figure*}

\begin{figure*}
  \centering
  \includegraphics[width=1.0\linewidth]{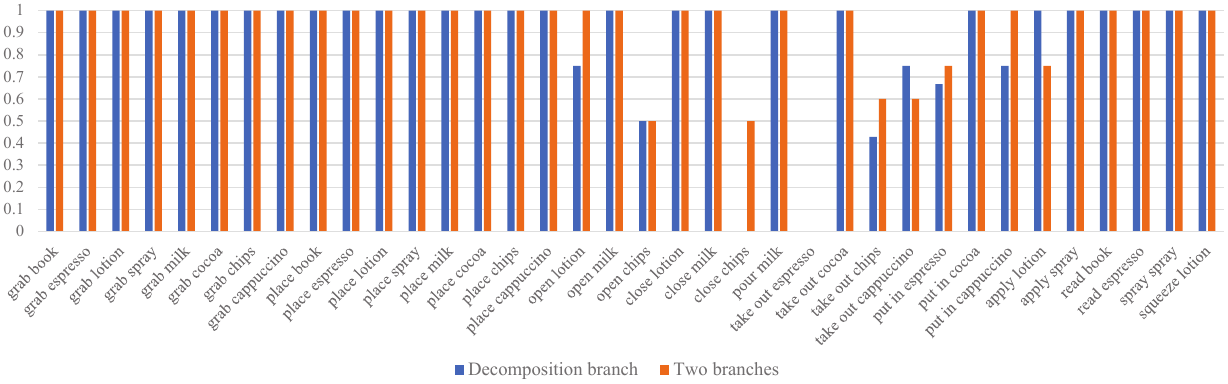}\\
  \caption{Results of each action class whether or not composing new training data on the H2O dataset.}
  \label{fig:H2Ocomp}
\end{figure*}

The proposed free-form composition networks mainly consist of two branches, i.e., a decomposition branch and a composition branch. To demonstrate the effectiveness of two branches for egocentric action recognition, we separately report the results of the decomposition branch and two branches together. The accuracies corresponding to two kinds of settings are shown in Table~\ref{tab:diffcomp}. On the something-something V2 dataset, the mean accuracy is $ 51.3\%$ based on the decomposition branch, and the result is increased to $ 52.9\%$ through generating new training data for the tail actions with the composition branch. We can see that some tail actions in this dataset can not be composed by the videos of other action classes. For instance, the verb ``cover" only appears in the action ``covering something with something", so we can not extract its visual feature from other actions. We thus gather 50 action classes (i.e., SS-V2-Subset) with shared verbs and prepositions/adverbs from something-something V2 to evaluate the effectiveness of two branches. The number of training samples in each action class is illustrated in Figure~\ref{fig:seleclas}, where the class label is consistent with original dataset. Based on the decomposition branch, the mean accuracy is $ 55.1\%$ using only the locations of hands and objects. Then, the accuracy is improved to $ 64.7\%$ by generating the features of new training samples for 20 action classes with the fewest training data. H2O is a relatively small dataset, and the number of videos for each class in the training set ranges from 12 to 25. We thus generate new training samples for all the action classes. After applying data augmentation, the mean accuracy is increased from $ 91.7\%$ with only the decomposition branch to $ 93.3\%$ with two branches together. The EPIC-KITCHENS-100 dataset expects the methods to separately predict the verb, noun, and action labels. It is worth noting that the two-branch architecture is designed to directly predict the action label. To test the effectiveness of the compositional branch, we only compare the action prediction accuracies whether or not generating new training data for the tail actions. The average accuracy is increased from $ 50.5\%$ to $ 52.2\%$ through data augmentation.

We also compare the performances of each action class whether or not composing new training samples. The comparison results on the SS-V2-Subset are shown in Figure~\ref{fig:SScomp}. The class label and the number of training samples in each class are both consistent with Figure~\ref{fig:seleclas}. With only the decomposition branch, none of the validation samples is correctly recognized in some tail classes, such as 136, 60, and 137. After composing new samples for 20 tail classes, the accuracies of 45 out of 50 actions are improved, and the result of action class 15 is increased from $ 50.0\%$ to $ 83.5\%$. We further compare the results of the decomposition and composition branches on the H2O dataset. As shown in Figure~\ref{fig:H2Ocomp}, the decomposition branch of our FFCN obtains promising results, and the accuracies of most classes are $ 100\%$. After composing new samples with two branches, the results of five classes are increased by a clear margin, and the accuracy of the action ``close chips" is increased from 0 to $ 50.0\%$. But some actions in this dataset are still easily confused with each other. For example, the action ``take out espresso" is usually misclassified into the actions of ``take out cappuccino" and ``put in espresso".

\section{Conclusion}
\label{section:Conclusion}

In this paper, we propose the free-form composition networks to relieve the data scarcity problem in the long-tailed and few-shot egocentric action recognition. The proposed free-form composition networks (FFCN) works in a flexible way for action recognition, regardless of the different numbers of verbs/prepositions/nouns used in the action description: it first extracts the spatial-temporal features of multiple different verbs, prepositions, and nouns appearing in the action description of videos, and then generates new training samples for the rare action classes by integrating disentangled verb, preposition, and noun feature representations together. Extensive experiments on popular egocentric action recognition datasets show the effectiveness of the proposed method to address the challenging data scarcity issues, containing the long-tailed and few-shot egocentric action recognition. In the future, we will investigate to adaptively learn the semantic components from the raw text, which is not restricted to the combination of verbs, prepositions, and nouns.



%

{
\bibliographystyle{IEEEtran}
\bibliography{egbib}
}

%
%
\begin{IEEEbiography}[{\includegraphics[width=1in,height=1.25in,clip,keepaspectratio]{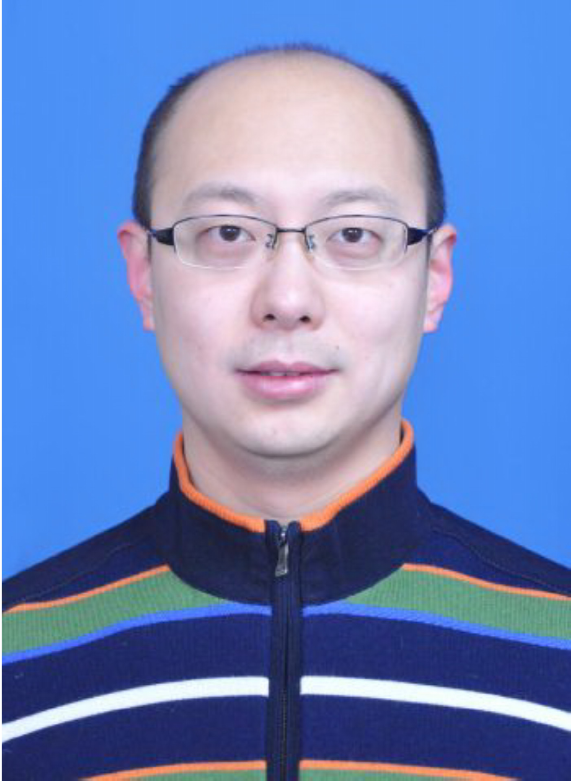}}]{Haoran Wang}
received the B.S. degree from the Department of Information Science and Technology, Northeastern University, China, in 2008, and the Ph.D. degree from the School of Automation, Southeast University, China, in 2015. In 2013, he was a visiting scholar at the Department of Computer Science of Temple University, USA. From 2018 to 2019, he was a visiting scholar at the School of Computer Science, University of Sydney. He is currently an associate professor at Northeastern University, China. His research interests include human action recognition, event detection, and machine learning.
\end{IEEEbiography}

\begin{IEEEbiography}[{\includegraphics[width=1in,height=1.25in,clip,keepaspectratio]{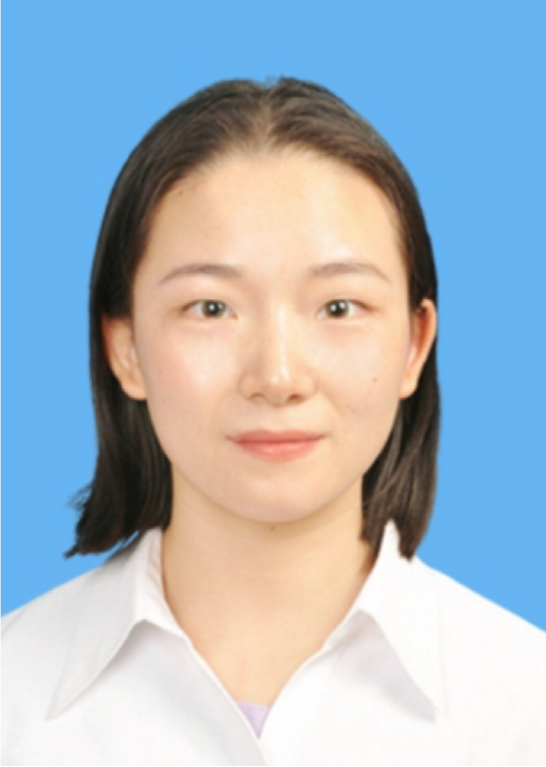}}]{Qinghua Cheng}
received the B.E. degree from the Department of Automation, Henan Polytechnic University, China, in 2021. She is currently pursuing the M.E. degree with Northeastern University, China. Her research interests include human action recognition, human-object interaction detection, and deep learning.
\end{IEEEbiography}

\begin{IEEEbiography}[{\includegraphics[width=1in,height=1.25in,clip,keepaspectratio]{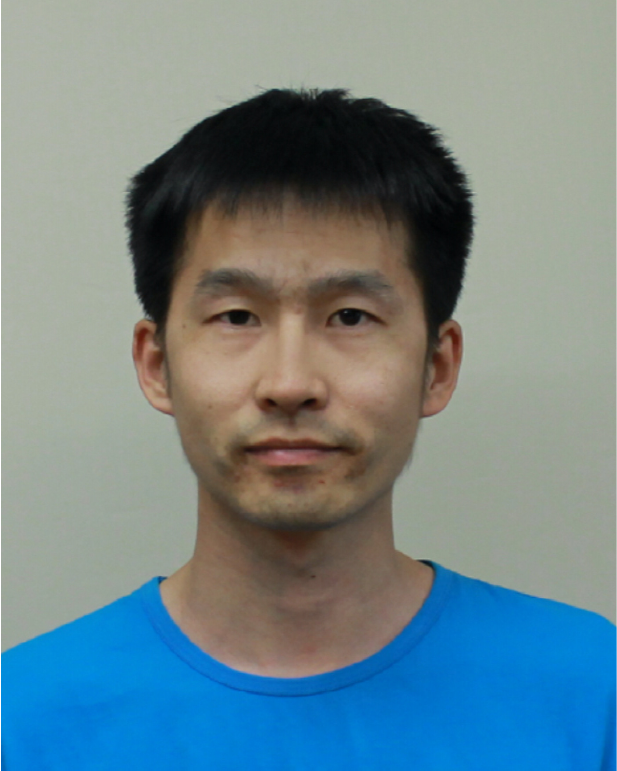}}]{Baosheng Yu}received a B.E. degree from the University of Science and Technology of China (USTC) in 2014, and a Ph.D. degree from the University of Sydney (USYD) in 2019. He is currently a Research Fellow in the School of Computer Science at the University of Sydney, Australia. His research interests include computer vision and machine learning. He has authored or co-authored over 40 publications on top-tier international conferences and journals, including CVPR, ICCV, ECCV, IJCV, and IEEE TPAMI.\end{IEEEbiography}

\begin{IEEEbiography}[{\includegraphics[width=1in,height=1.25in,clip,keepaspectratio]{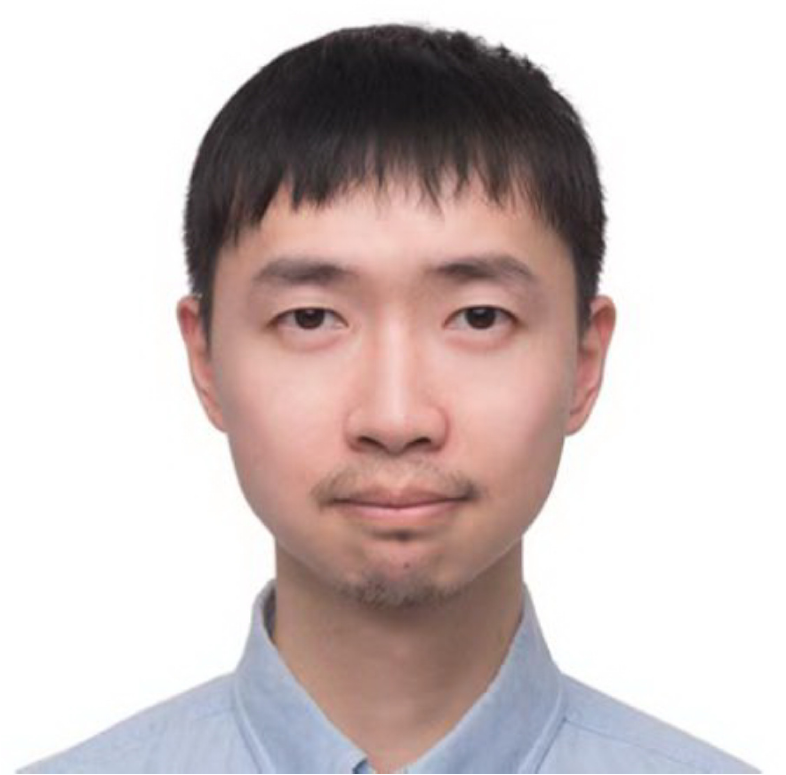}}]{Yibing Zhan}
obtained his bachelor's degree and doctor's degree from the school of information science and technology at the University of Science and Technology of China in 2012 and 2018. After graduating with a doctor's degree, from 2018 to 2020, Yibing Zhan served as an associate researcher in the school of computer science of Hangzhou Dianzi University. Now, Yibing Zhan works in the JD Explore Academy as an algorithm scientist. He mainly explores scene graph generation, foundation model, and graph neural networks. He has published many scientific papers in top conferences and journals such as NeurIPS, CVPR, ACM MM, ICCV, and IEEE TMM.
\end{IEEEbiography}

\begin{IEEEbiography}[{\includegraphics[width=1in,height=1.25in,clip,keepaspectratio]{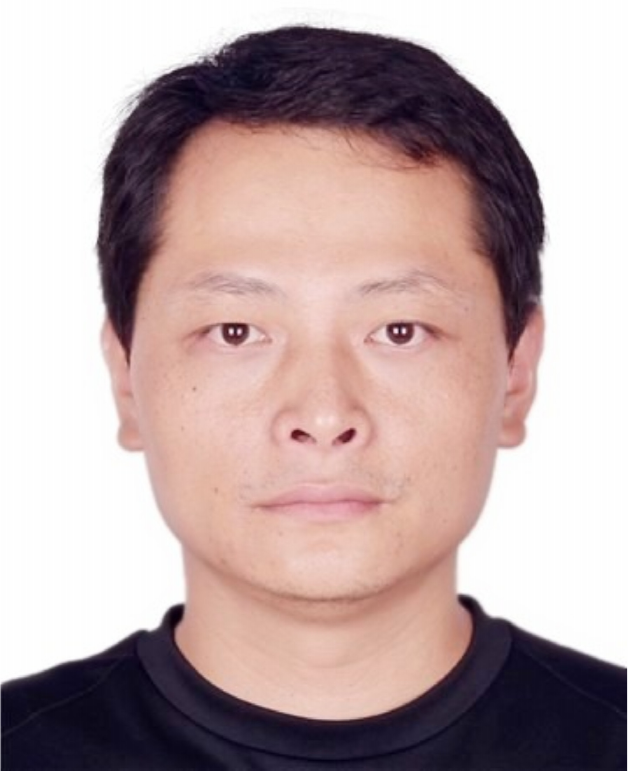}}]{Dapeng Tao}
received the B.E. degree from Northwestern Polytechnical University, Xi¡¯an, China, in 1999, and the Ph.D. degree from the South China University of Technology, Guangzhou, China, in 2014. He is currently a Professor with the School of Information Science and Engineering, Yunnan University, Kunming, China. He has authored or co-authored over 100 scientific articles at top venues, including IEEE TIP, IEEE TNNLS, IEEE TCYB, IEEE TMM, IEEE CSVT, Pattern Recognition, CVPR, ICCV, IJCAI, AAAI.
\end{IEEEbiography}

\begin{IEEEbiography}[{\includegraphics[width=1in,height=1.25in,clip,keepaspectratio]{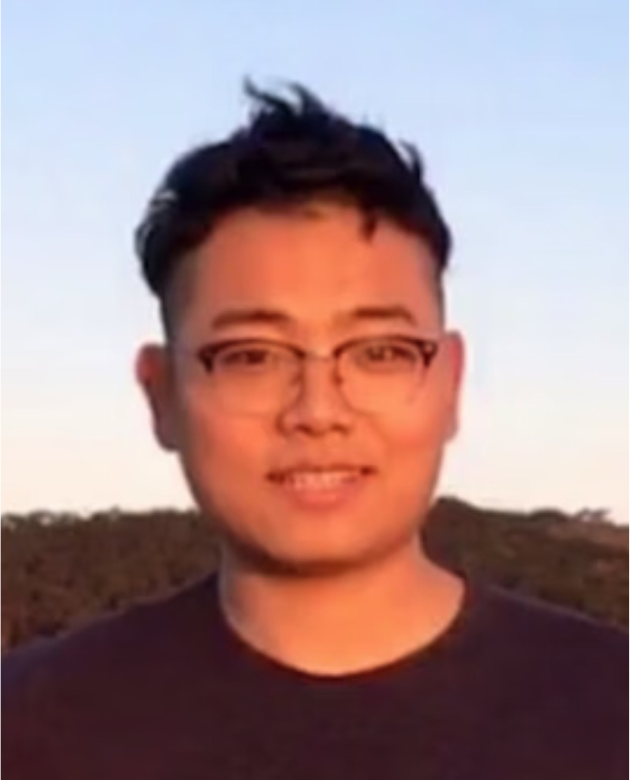}}]{Liang Ding}
received Ph.D. from the University of Sydney. He is currently an algorithm scientist with JD.com and leading the NLP research group at JD Explore Academy. He works on deep learning for NLP, including language model pretraining, language understanding, generation, and translation. He published over 30 research papers in NLP/AI, including IEEE T-KDE, IEEE T-MM, ACL, EMNLP, NAACL, COLING, ICLR, AAAI, SIGIR, and CVPR, and importantly, some of his works were successfully applied to the industry. He served as Area Chair for ACL 2022 and Session Chair for SDM 2021 and AAAI 2023. He won many AI challenges, including SuperGLUE/ GLUE, WMT2022, IWSLT 2021, WMT 2020, and WMT 2019.
\end{IEEEbiography}

\begin{IEEEbiography}[{\includegraphics[width=1in,height=1.25in,clip,keepaspectratio]{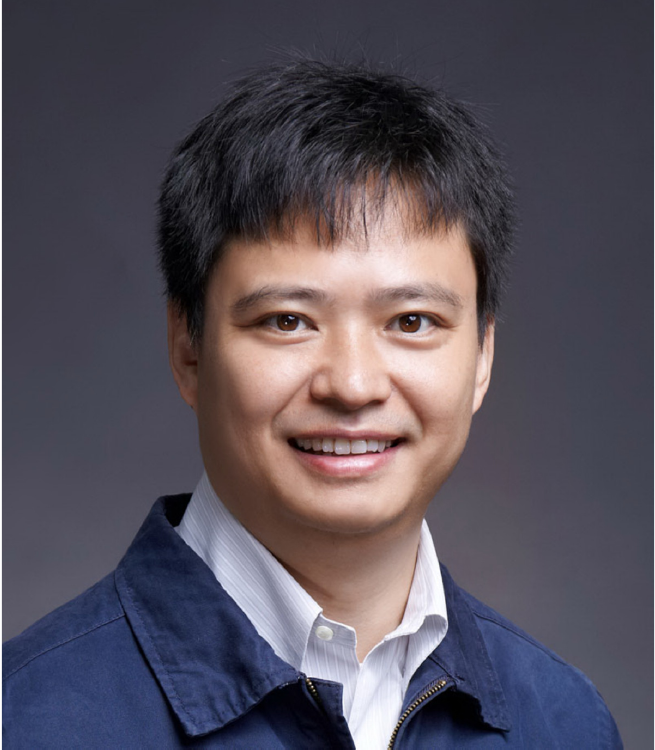}}]{Haibin Ling}
received B.S. and M.S. from Peking University in 1997 and 2000, respectively, and Ph.D. from University of Maryland in 2006. From 2000 to 2001, he was an assistant researcher at Microsoft Research Asia; from 2006 to 2007, he worked as a postdoctoral scientist at UCLA; from 2007 to 2008, he worked for Siemens Corporate Research as a research scientist; and from 2008 to 2019, he was a faculty member of the Department of Computer Sciences for Temple University. In fall 2019, he joined the Department of Computer Science of Stony Brook University, where he is now a SUNY Empire Innovation Professor. His research interests include computer vision, augmented reality, medical image analysis, visual privacy protection, and human computer interaction. He received Best Student Paper Award of ACM UIST (2003), Best Journal Paper Award at IEEE VR (2021), NSF CAREER Award (2014), Yahoo Faculty Research and Engagement Award (2019), and Amazon Machine Learning Research Award (2019). He serves or served as associate editors for IEEE Trans. on Pattern Analysis and Machine Intelligence (PAMI), IEEE Trans. on Visualization and Computer Graphics (TVCG), Pattern Recognition (PR), and Computer Vision and Image Understanding (CVIU), and as Area Chairs for major computer vision conferences such as CVPR, ICCV, ECCV and WACV.
\end{IEEEbiography}

%

\vfill

\end{document}